Bio-inspired Dual-auger Self-burrowing Robots in Granular Media

by

Md Ragib Shaharear

A Thesis Presented in Partial Fulfillment
of the Requirements for the Degree
Master of Science

Approved April 2023 by the
Graduate Supervisory Committee:

Junliang (Julian) Tao, Chair
Edward Kavazanjian
Hamidreza Marvi

ARIZONA STATE UNIVERSITY

May 2023

# ABSTRACT

It has been found that certain biological organisms, such as *Erodium seeds* and *Scincus scincus*, are capable of effectively and efficiently burying themselves in soil. Biological Organisms employ various locomotion modes, including coiling and uncoiling motions, asymmetric body twisting, and undulating movements that generate motion waves. The coiling-uncoiling motion drives a seed awn to bury itself like a corkscrew, while sandfish skinks use undulatory swimming, which can be thought of as a 2D version of helical motion. Studying burrowing behavior aims to understand how animals navigate underground, whether in their natural burrows or underground habitats, and to implement this knowledge in solving geotechnical penetration problems. Underground horizontal burrowing is challenging due to overcoming the resistance of interaction forces of granular media to move forward. Inspired by the burrowing behavior of seed-awn and sandfish skink, a horizontal self-burrowing robot is developed. The robot is driven by two augers and stabilized by a fin structure. The robot's burrowing behavior is studied in a laboratory setting. It is found that rotation and propulsive motion along the axis of the auger's helical shape significantly reduce granular media's resistance against horizontal penetration by breaking kinematic symmetry or granular media boundary. Additional thrusting and dragging tests were performed to examine the propulsive and resistive forces and unify the observed burrowing behaviors. The tests revealed that the rotation of an auger not only reduces the resistive force and generates a propulsive force, which is influenced by the auger geometry, rotational speed, and direction. As a result, the burrowing behavior of the robot can be predicted using the geometry-rotation-force relations.

# DEDICATION

This thesis is dedicated to my mother and father.

ACKNOWLEDGMENTS

I would like to express my sincere gratitude to my advisor, Dr. Junliang (Julian) Tao, at Arizona State University for his excellent guidance and mentorship throughout my MS research work. Without his invaluable suggestions and ideas, this thesis would not have been possible. His wealth of knowledge has been invaluable in developing me as a researcher. I am also grateful to my thesis committee members, Dr. Edward Kavazanjian and Dr. Harmidreza Marvi, for their invaluable suggestions and advice on this study. Additionally, I am thankful to all of my co-workers Yong Tang, Yi Zhong, Xiwei Li, Sarina Shahhosseini, and Mohan Parekh for giving valuable suggestions throughout my MS degree. Lastly, I want to express my gratitude to my beloved wife, Ismat Jahan Zinia, for her continuous support in completing my MS degree.

TABLE OF CONTENTS

<208
<208

<208

<208

LIST OF TABLES

LIST OF FIGURES

Figure                                                                                                                  Page

Chapter 1

INTRODUCTION

1.1 Motivation and Objective

Penetration problem is a common engineering problem in Geotechnical engineering. Geotechnical engineering often faces challenges involving soil displacement in a vertical or horizontal direction using items like piles, tunnels, or penetrometers. These horizontal and vertical penetration problems have been extensively investigated using numerical simulation methods and physical experiments. All vertical and horizontal penetration problems involve external forces or platforms, while our objective is to achieve self-burrowing.

In Geotechnical engineering, one of the primary challenges is achieving self-burrowing capability without relying on external platforms or forces. Self-burrowing is the ability to move through soil without relying on external devices or machines. The objective of this thesis is to achieve self-burrowing. Self-burrowing is a novel and bio-inspired area of research that involves creating robots or tools capable of moving independently through soil by mimicking the burrowing strategies of living organisms such as earthworms, razor clams, or Erodium seeds. However, self-penetration still needs to be well investigated because modelling and comprehending the interactions between the soil and the robot is challenging. Therefore Self-burrowing is a new and interdisciplinary research topic that requires more innovation and further research. Self-burrowing robots have potential applications in potential geotechnical site investigation, contamination detection, precision agriculture, and extraterrestrial exploration.

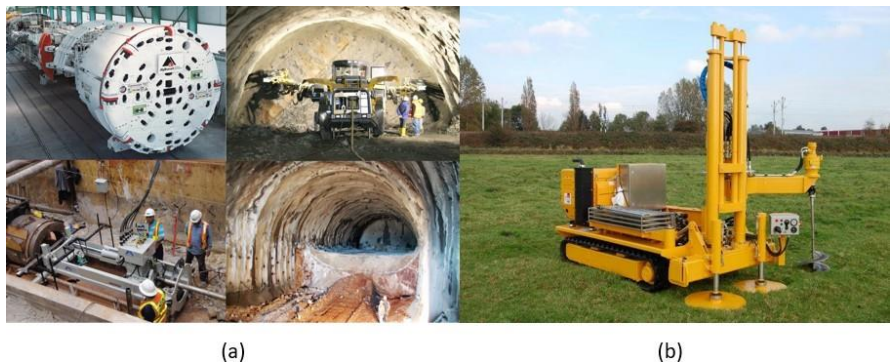

**Figure 1.1:** Illustration of various methods of geotechnical penetration : (a) Tunnel boring machines (TBMs), drill and blast, shielded excavation machines (SEMs), and micro tunnel boring machines (MTBMs). The four methods are illustrated in clockwise order from the top left corner of the figure. The images used in the illustration are obtained from sources such as Heitkampt-swiss.ch, Akkerman.com, and a publication by Hoek et al. in 2007. ; (b) Cone Penetration Test(CPT)- Image from Structville

## 1.2 Background and Previous Study

Robotics have various applications in sensing, autonomous locomotion on land, underwater swimming, and flying in the air (Siciliano and Khatib, 2016). However, developing an autonomous underground robot capable of moving into granular media is challenging. Burrowing into the granular media associated with soil diversity, interactions between soil entities, and distortion. However, observing the efficient and effective locomotion of biological organisms within granular media can provide valuable inspiration for addressing the challenges of self-burrowing. Biological organisms can survive underground by managing different burrowing processes. For example, Sandfish Shink (*S. scincus*) can locomote into the granular media by undulatory swimming strategy which is a 2D version of helical motion (Maladen *et al.*, 2009), earthworms can move into the ground by the action of muscles in the body wall (Dorgan, 2015), bivalve mollusks use dual-anchor strategy (Trueman, 1975), plant roots expand through the soil by 'tip extension' process (Abdalla *et al.*, 1969).

This thesis is particularly inspired by the burrowing mechanism of *Erodium seeds* for rotation and *sandfish skink* for undulatory swimming, which is a 2D version of helical motion. *Erodium seeds* and *sandfish skink* have an exceptional morphology that helps them to self-burial into the granular media. When seeds come into contact with the ground, they exhibit a coiling and uncoiling motion, which enables them to burrow downward. This motion is facilitated by their moisture-responsive awns, which have a helical shape (Jung *et al.*, 2017a).

## 1.3 Problem Statement

The burrowing behavior of *Erodium seeds* for rotation, *sandfish skink* for undulatory swimming, which is a 2D version of helical motion, and *sphylum spirochaetes* for rotation in a low-Reynold number environment is a major area of research to learn burrowing dynamics. This thesis also observes that many biological organisms follow different strategies and diverse traits based on the surrounding environment and soil conditions. These different strategies and diverse traits depend on burrowing kinematics, morphology and properties of the medium. They achieve the burrowing by increasing the anchorage or by reducing the resistance. In granular media symmetry, breaking can be achieved by rotational penetration (Tang *et al.*, 2020a; Jung *et al.*, 2017b) and reduction of granular drag are also important to achieve self-burrowing. To achieve the autonomous self-burrowing behavior thrust and drag forces generated by robots need to be studied to understand the burrowing behavior better. Therefore, further research on burrowing strategies is necessary to transfer the biological organisms burrowing behaviors into engineering applications.

## 1.4 Research Goal

The final goal of this study is to understand the burrowing behavior of biological organisms (with the emphasis of *Erodium seeds* and *sandfish skink*) in the glass beads and Ottawa sand F65 granular media to create a self-burrowing technology inspired by biological organisms for exploring underground environments. The goals of this study are:

- To investigate common burrowing strategies in nature and common penetration problems to achieve self-burrowing;

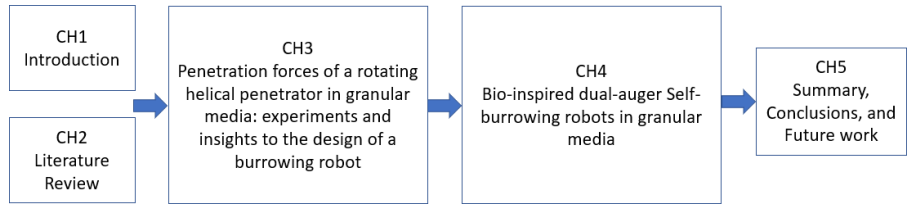

**Figure 1.2:** Outline of the Thesis

- Penetration forces of a rotating helical penetrator in granular media: experiments and insights to the design of a burrowing robot;

- To investigate preliminarily, how a rotating right-handed auger facilitates upward and downward penetration without any external forces;

- Bio-inspired dual-auger Self-burrowing robots in granular media;

## 1.5   Organization of the Thesis

This thesis is structured into five chapters, and the outline is illustrated in figure 1.2.

The first chapter describes how the burrowing behavior of *Erodium seeds sandfish skink* served as the basis for developing a self-burrowing robot. In the first chapter, four primary aims are proposed, and each of these objectives is subsequently discussed in the following chapters.

In chapter two, the literature review is discussed. The literature review consists of four major parts. A comprehensive overview of the burrowing behavior of *Erodium seeds* and *sandfish skink* is discussed in section 2.1, Existing burrowing mechanisms of the bio-inspired self-burrowing robot are discussed in section 2.2, Diverse methods of burrowing in nature is discussed in section 2.3 and Scallop theorem and why symmetry breaking is important is discussed in section 2.4.

In chapter three, Penetration forces of a rotating helical penetrator in granular media: experiments and insights to the design of a burrowing robot are discussed, and how a rotat-ing auger can facilitate rotational penetration is investigated.

Chapter four of this research project is dedicated to assessing the self-burrowing potential of bio-inspired dual-auger robots in granular media. The hypothesis that self-burrowing requires a thrust force greater than or equal to the drag force is introduced and analyzed.

Chapter five summarises the key findings in the thesis and offers suggestions for further research.

Chapter 2

LITERATURE REVIEW

In this chapter, the literature review is divided into four major sections (1) Previous work on burrowing behavior of *Erodium seeds* and *Scincus scincus* (2) Existing burrowing mechanisms of a bio-inspired self-burrowing robot (3) Diverse methods of burrowing in nature (4) Scallop theorem and symmetry breaking.

## 2.1 Burrowing Behavior of *Erodium seeds* and *Scincus scincus*

Underground locomotion is challenging due to the substrate's dense and complex soil behavior. However, many biological organisms adapted to locomote effectively and efficiently through the underground to find food and survive. The burrowing robotics community borrow those burrowing strategies and designs a series of robots to solve many engineering problems. Locomotion through granular media is observed for different biological animals (Maladen *et al.*, 2009; Jung, 2010; Hosoi and Goldman, 2015; Li *et al.*, 2021) and flowering plants (Bengough *et al.*, 2008; Sadeghi *et al.*, 2017) by reducing penetration resistance. It is observed that some flowering plants such as *Erodium seeds* and *Pelargonium species* have shown self-burial behavior for hygroscopic coiling and uncoiling of the awns (Evangelista *et al.*, 2011). They belong to the genus Geranium family, and their produced seed is used for dispersal and burial (Jung *et al.*, 2014; Evangelista *et al.*, 2011; Stamp, 1984). Some other seeds also spread from the parent plant and can be buried for germination and survival. Their self-burial feature can be achieved by hygroscopic expansiveness awns (Abraham *et al.*, 2012; Geer *et al.*, 2020), drilling into the soil, twisting and untwisting with respect to changing in humidity, and modified hygroscopically powered helical shape. The rotational penetration movement caused by this cyclic process enables the *erodium seeds* to bury themselves into the soil, facilitating its future germination. *Scincus scincus* show an Undulation swimming movement when moving on the surface, which is a 2D version of helical motion (Maladen *et al.*, 2009).

## 2.2 Existing Burrowing Mechanisms of Bio-inspired Burrowing Robots

Recently many researchers have shown interest in converting biological burrowing mechanisms into engineering solutions. For example, underground investigation systems are inspired by the burrowing behavior of worms, underground drilling systems are inspired by hygroscopically powered helical shape awn, and inspired by razor clam by reducing penetration resistance using fluidization (Huang, 2020). These burrowing behavior help us to clear our understanding of biological burrowing (Aguilar *et al.*, 2016) and generate ideas to convert biological burrowing into engineering solutions. Worm-inspired burrowing robots follow a dual-anchor burrowing process for burrowing. An inchworm-like drilling system is proposed by (Rafeek *et al.*, 2001) for subterranean exploration. To explore the seafloor, an earthworm-inspired robot is developed by Isaka *et al.* (2019). (Wright *et al.*, 2007) proposed a snake-inspired robotic design for search and rescue purposes. A fish-inspired robot is studied to improve the design of autonomous underwater vehicles (Kodati and Deng, 2009). Bio-inspired flying robot is developed to explore the autonomous indoor flying challenges (Zufferey, 2008). Bio-inspired flying robots have been investigated to improve the agility and durability of indoor flight (Zufferey, 2008).

Different types of underground burrowing robots burrowing strategies are shown in table 2.1.

## 2.3 Diverse Methods of Burrowing in Nature

Different kinds of underground burrowing strategies are found in nature for different biological organisms. Through these subterranean burrowing techniques, these living bi-ological organisms can create adequate anchorage and thrust to surmount the resistance presented by the adjacent soil, ultimately allowing them to make subterranean progress for their survival and to find food. Dual-anchor burrowing strategies are found for biological organisms such as bivalve mollusks (Koller-Hodac *et al.*, 2010), stony coral (Kohls *et al.*, 2021), *Polychaete* (Pak *et al.*, 2006), and parasitoid wasps (Harvey *et al.*, 2003). Plant root shows Tip extension burrowing strategy (Mazzolai *et al.*, 2011). *Labridae* fish follow the Diving/Plunging burrowing strategy to locomote through the underground. Seed awns show rotational drilling for underground penetration (Evangelista *et al.*, 2011). Different biological living organisms may have evolved similar traits because they faced similar chal-lenges. On the other hand, organisms of the same kind may have different traits to cope with different environments and situations. These two features can be termed convergent and divergent evolution (Rachna, 2017). The traits that have changed to be more alike or different are usually about shape and movement. These traits are designed to facilitate underground locomotion and burrowing by either improving anchorage or decreasing the resistance of the surrounding soil. Bivalve clams organisms show the traits of shell rocking (Trueman *et al.*, 1966) and shell penetration (Trueman, 1967). *Polychaete* shows inner bodily enlargement and wider head oscillation traits during underground movement (Francoeur and Dorgan, 2014). Burrowing fish facilitate underground burrowing by head-first penetration, and tail-first penetration (Herrel *et al.*, 2011). *Erodium seeds* achieve self-burrowing by twisting and untwisting characteristics (Evangelista *et al.*, 2011; Elbaum *et al.*, 2007).

**Table 2.1:** Different Types of Underground Burrowing Strategies

| Burrowing Strategy | Biological Organisms | References |
|---|---|---|
| Dual-anchor | bivalve mollusks | (Koller-Hodac *et al.*, 2010) |
| | stony coral | (Kohls *et al.*, 2021) |
| | *Polychaete* | (Pak *et al.*, 2006) |
| | parasitoid wasps | (Harvey *et al.*, 2003) |
| Undulation | sand eel | (Chen *et al.*, 2021) |
| | fossorial (burrowing) snakes | (Spranklin, 2006; Hopkins *et al.*, 2009) |
| | burrowing eel | (La Spina *et al.*, 2007) |
| | sandfish skink | (Šmíd *et al.*, 2021) |
| Cyclic rotation | *Phylum Spirochaetes* | (Kulkarni *et al.*, 2022) |
| | *Erodium Seeds* | (Fiorello *et al.*, 2022) |
| Tip expansion | plant root | (Mazzolai *et al.*, 2011) |
| Local fluidization | razor clam | (Tao *et al.*, 2020) |

## 2.4 Scallop Theorem and Symmetry Breaking

American physicist Edward Mills Purcell demonstrated that Swimming through a reciprocating motion in a low Reynolds number environment would cause an organism to have zero net translation (Purcell, 1977). Here the Reynold number is (Wikipedia, 2023):

$$\text{Reynold number} = \frac{\text{Inertial forces}}{\text{Viscous forces}}, \tag{2.1}$$

$$Re = \frac{\rho u L}{\mu} \tag{2.2}$$

The Reynolds number can be defined as the ratio of inertial forces to viscous forces within a fluid. When the value of the Reynolds number is low, the viscous forces dominate, resulting in a laminar and smooth flow (Wikipedia, 2023). The Reynolds number is represented by $Re$. It is calculated using the following parameters: $\rho$, which is the density of the fluid measured in kg/m$\hat{3}$, $u$, which denotes the velocity of the fluid in m/s, $L$, which is a characteristic length of the object, such as the diameter of a pipe or the length of an object, in meters, and $\mu$, the dynamic viscosity of the fluid in Pa*s (Pascal-seconds) (Wikipedia, 2023).

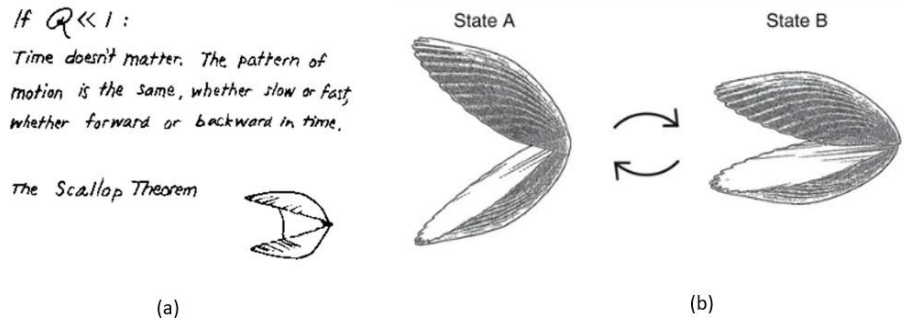

**Figure 2.1:** Illustration of Scallop Theorem: (a) Scallop Theorem Demonstration from Purcell Paper (Purcell, 1977).; (b) Purcell's Swimming Scallop Opens and Closes its Shell- Image from Wikipedia

Here the reciprocal motion is a type of movement pattern where the body shape changes in a way that is symmetrically repeated over time. This demonstration can be called the "scallop theorem": When a scallop performs a one-degree-of-freedom motion of opening and closing its shell in a fluid with a low Reynolds number environment, its movement kinematics will be reciprocal. This implies that the scallop will return to its original position after completing one motion cycle. Due to the symmetrical drag forces and propulsive effects resulting from its scallop movement, the scallop can retrace its steps exactly and return to its starting position.

In order to achieve net translation when swimming in a low-Reynolds number fluids environment, the swimmer's movement kinematics must be characterized by asymmetry, non-reciprocity or non-time-reversal. Burrowing in dry granular media can be compared to swimming in a low-Reynolds number fluids environment because both are inertialess and affected by resistance forces (Hosoi and Goldman, 2015). In the Purcell paper (Purcell, 1977), three different asymmetry kinematics are discussed to break the kinematics symmetry so that organisms can swim in the low Reynold number environment. Three kinds of

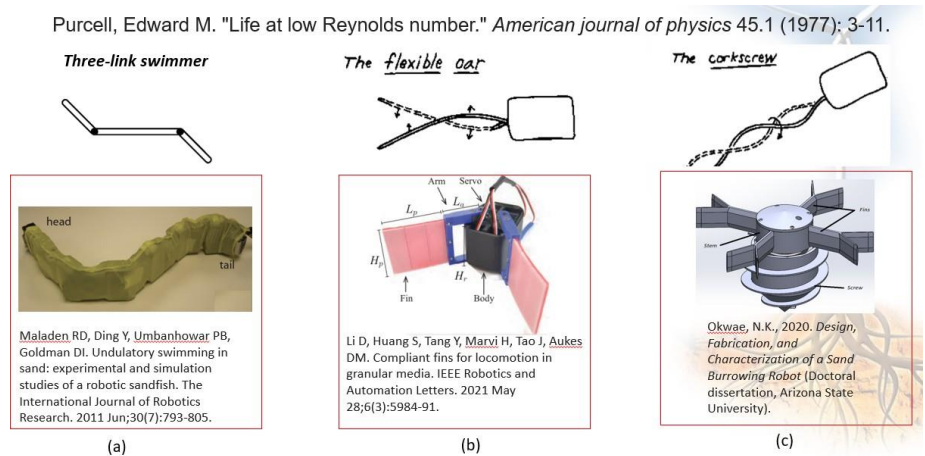

**Figure 2.2:** Illustration of Purcell Proposed(Purcell, 1977) Asymmetry Kinematics and Burrowing Robotics Community Robot Design : (a) Three Link swimmer Inspired Undulatory Swimming Robot by Professor Daniel Goldman Group from Georgia Tech (Maladen *et al.*, 2011a).; (b) The Flexible Oar Inspired Sand Swimming Robot Developed by Professor Daniel Aukes Group from ASU (Li *et al.*, 2021) (c) Corkscrew Inspired Sand Burrowing Robot Developed by Professor Hamidreza Marvi Group from ASU (Okwae, 2020)

.

asymmetric kinematics are the three-link swimmer, the flexible Oar and the corkscrew, as shown in the upper part of the figure 2.2. The burrowing robotics community also borrow these ideas and designs a series of burrowing robots.

For example, Inspired by three link swimmer, an undulatory swimming robot is developed by Professor Daniel Goldman's group from Georgia Tech (Maladen *et al.*, 2011a) as shown in figure 2.2 (a), the flexible oar-inspired sand swimming robot is developed by our collaborator Professor Daniel Aukes group from ASU (Li *et al.*, 2021) as shown in figure 2.2 (b). Corkscrew design is well studied, and a sand burrowing robot is developed by Professor Hamidreza Marvi's Group from ASU (Okwae, 2020) as shown in figure 2.2 (c). The sand-burrowing robot features a screw with fins that serve as an anchor, facilitating its movement through the medium, and the robot can move vertically in the upward and downward direction (Okwae, 2020). Our research goal is to design horizontal dual-auger self-burrowing robots so that we can control the burrowing direction from left to right and right to left. Initially, our investigation focused on how a rotating right-handed auger can facilitate rotational downward penetration (Shaharear and Tao, 2023; Shaharear *et al.*, 2023). Subsequently, we developed a dual-auger self-burrowing robot capable of horizontal movement through underground environments.

Chapter 3

PENETRATION FORCES OF A ROTATING HELICAL PENETRATOR IN GRANULAR MEDIA

3.1 INTRODUCTION

Moving into the soil is challenging due to the intrinsic gravitational field, which causes effective stress and soil shear strength to increase with depth. However, many burrowing species live underground and have well-developed movement strategies. The movement mechanism into the soil relies upon the properties of the medium and size of organisms (Dorgan, 2015).

For example, a sand-swimming lizard, an undulatory sandfish, can swim within granular media (Maladen *et al.*, 2011b). Earthworms and mole cricket can dig up the soil in front of them and pull it back so they can go forward (Moon *et al.*, 2013). Bivalves can burrow themselves into the sediment using the two-anchor process (Koller-Hodac *et al.*, 2010). The Atlantic razor clam (*Ensis directus*) can burrow into the soil by contracting its valves, which causes the surrounding soil to fluidize and reduces burrowing drag (Dorsch and Winter, 2015). There are two types of *Erodium seeds*: one is capable of autochory, which allows it to disperse the seeds away from the parent plant, and the other type adopts a self-burial strategy to bury itself in the soil. The capability of these movements is attributed to the presence of hygroscopic tissues that generate passive motion by altering the hydration levels of the cell walls. The self-burial process is facilitated by a unique dispersal unit (the awn) present in each seed. This awn responds to changes in environmental humidity by altering its shape. It adopts a helical shape when dry, whereas when wet, it becomes linear (Stefano Mancuso, Barbara Mazzolai, Diego Comparini, Liyana Popova,

Elisa Azzarello, Elisa Masi, Nadia Bazihizina, Edoardo Sinibaldi, 2014). *Erodium seeds* move across the surface and into the ground thanks to coiling and uncoiling motor action caused by the day-night humidity cycle (Stamp, 1984) (Elbaum *et al.*, 2007). These species can be biological models for a burrowing robot that can automatically travel through soil.

Internal forces and body twists allow burrowing organisms to move. However, organisms coordinate the movement of several body parts to improve generations of anchoring and propulsion, allowing them to resist backward slip and march forward. Similarly, organisms alter the morphology of numerous body parts to improve locomotor efficacy and efficiency. Underground locomotion in nature is essentially a problem of soil-organism interaction, similar to the challenges of soil-structure interaction in geotechnical engineering (Huang and Tao, 2020). This study, motivated by a biological burrowing mechanism of *Erodium seeds* and *sandfish skink*, performs a series of downward rotational burrowing tests to preliminarily evaluate the burrowing performance of helical penetrator under different conditions. Propulsive force is induced along the axis of the rotating helical penetrator, which facilitates symmetry breakdown and reduces anisotropic frictional forces (Darbois Texier *et al.*, 2017). The findings of these experiments have implications for the future creation of a unique self-burrowing two-auger robot that can burrow in different directions and can be used for geotechnical site investigation, underground sensing, precision agriculture, and other applications.

## 3.2 METHODOLOGY

### *3.2.1 Clockwise and Counterclockwise Penetration Tests*

It has been proven that breaking kinematic symmetry or granular media boundary conditions can result in net translations (Tao *et al.*, 2020) (Maladen *et al.*, 2011b). It has been found that rotation reduces penetration resistance (Tang and Julian Tao, 2021) (Tang *et al.*,

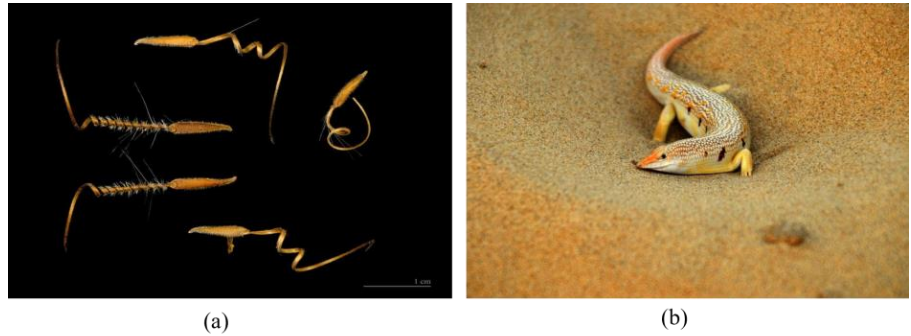

**Figure 3.1:** Bio-inspired Models from Nature: (a) *Erodium Seeds* awn; (b) *Sandfish Skink* (3.1 (a): Image Courtesy of the Attribution-Share Alike 4.0 International; (3.1 (b): Image Courtesy from through the Sandglass Magazine).

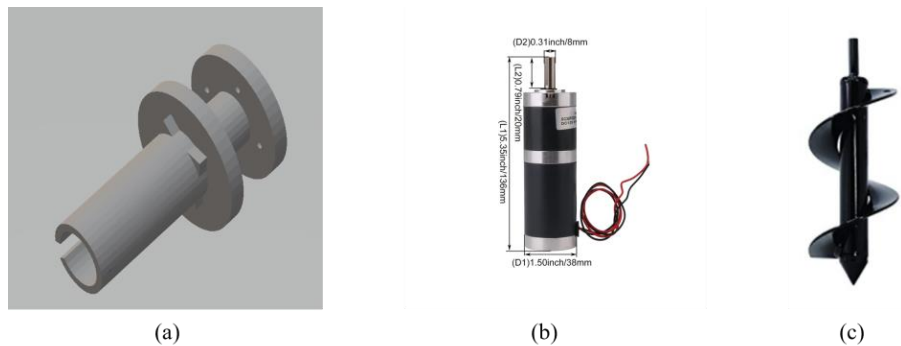

**Figure 3.2:** Different Parts for Experiment: (a) Universal Robot (UR16e) Connector and DC motor Holder ; (b) DC Motor :  (c) Auger (3.2 (a):  Autodesk Inventor Designed 3D Model; 3.2 (b): Commercial Product from Amazon; 3.2 (c): Commercial Product from Amazon )

2020b). By coordinating rotational motion and asymmetry shape, it is possible to break the symmetry of the kinematics of soil.

In this study, the penetration forces of a right-handed rotating helical penetrator are explored through laboratory tests in granular media. The rotating helical penetrator consists of two major segments: a 12V DC rotational gear motor (Size:136mmx38mm) whose speed range is 50 revolutions per minute (RPM) and a helical auger (size:177.8mmx50.8mm). Ottawa sand F65 ($D_{50}$ = 0.2 mm) and glass beads ($D$ = 3.0 mm) are used as granular media. Two cylindrical buckets with a diameter of 28 cm and a depth of 36 cm are used for sample preparation. The soil sample preparation method is important for conducting any experimental laboratory test. However, to keep the density of the samples accurate and precisely dry pluviation preparation technique is applied (Okamoto and Fityus, 2017). The full experimental setup consists of a universal robotic arm, a 3D printed motorized coupler as shown in figure 3.3 (a), a powerful high torque DC motor (Rated torque: 19.5 kg.cm, Rated current: 2.1A, Power: 25W) as shown in figure 3.2 (b), and a helical penetrator as shown in figure 3.2 (c). The rotational penetration tests are conducted with the aid of the UR16e universal robot. The universal robotic arm can measure the penetration force and torque, considering real-time feedback from the process during penetration. It includes six-axis force and torque sensor systems that capture and convert mechanical loads into forces and torques along all axes. The recorded penetration force and torque data from the robotic arm's control box were sent to the local drive using the Real-Time Data Exchange Python package. In this experiment, the robotic arm is used to move vertically. DC gear motor facilitates rotational penetration in a clockwise and counterclockwise manner considering a right-handed helical penetrator as shown in figure 3.3 (b) and figure 3.3 (c). An ATmega2560 microcontroller controls the rotational movement of the DC gear motor. Arduino ATmega2560 microcontroller(Islam *et al.*, 2018; Prayash *et al.*, 2019; Shaharear *et al.*, 2019) can generate a PWM signal to facilitate clockwise and counterclockwise rotation. This study is under different rotational (10 rpm, 30 rpm, 40 rpm, 45 rpm, 50 rpm) and vertical (0.04 m/s, 0.01 m/s, 0.001 m/s) velocities, downward penetration tests are conducted. All tests are conducted three times to make sure the experiment data is identical for every test.

### 3.2.2  Sliding Test

Figure 3.4 illustrates the sliding test of a right-handed rotating auger. The sliding test is designed to investigate the burrowing behavior of a right-handed rotating auger without any external effect. The sliding test consists of a slider and a rotating helical auger.

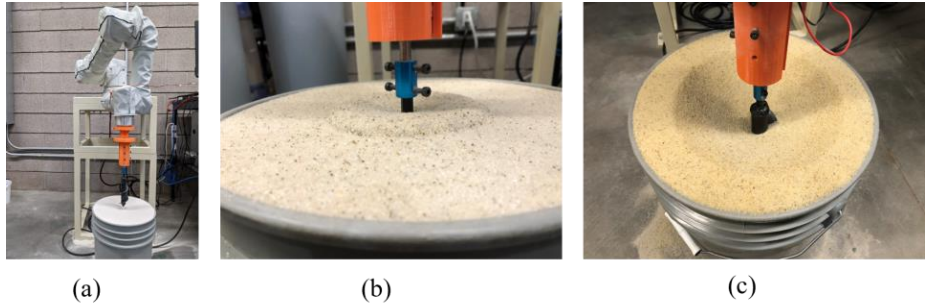

**Figure 3.3:** Illustration of Testing Setup: (a) Full Experiment Setup with Universal Robot (UR16e) and Motorized Helical Penetrator; (b) Clockwise Penetration Test : (c) Counter-clockwise Penetration Test

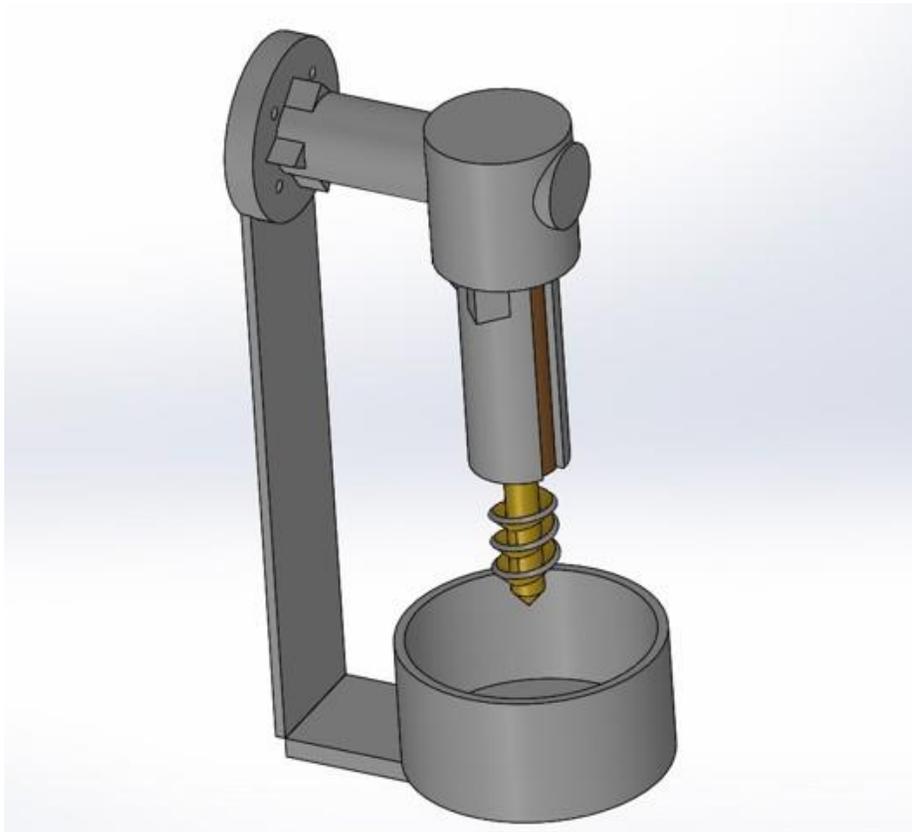

**Figure 3.4:** Illustration of Sliding Test Setup: Sliding Test Setup Consists of a Slider and Right-handed Rotating Auger

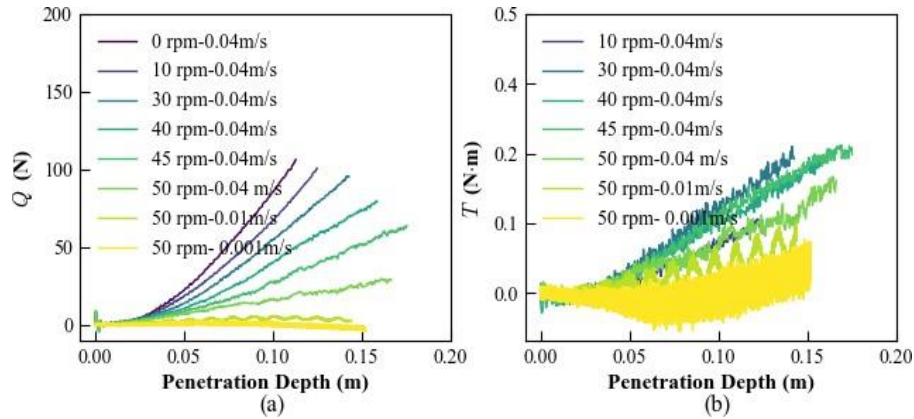

**Figure 3.5:** Clockwise Penetration Force and Torque for Helical Penetrator Under Different Rotational Velocities in Ottawa Sand F65: (a) Penetration Force ($Q$); (b) Penetration Torque ($T$).

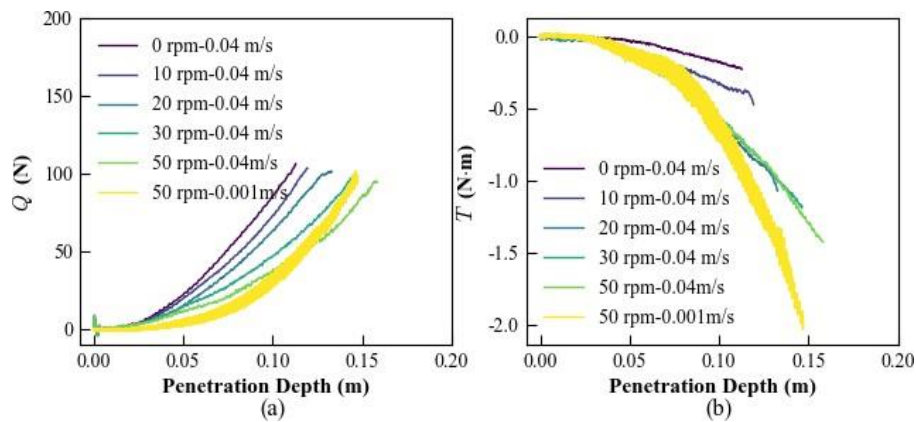

**Figure 3.6:** Counterclockwise Penetration Force and Torque for Helical Penetrator Under Different Rotational Velocities in Ottawa Sand F65: (a) Penetration Force ($Q$); (b) Penetration Torque ($T$).

## 3.3 RESULTS AND DISCUSSIONS

### 3.3.1 *Clockwise and Counterclockwise Downward Penetration in Ottawa Sand F65*

Different clockwise and counterclockwise penetration forces and torque in Ottawa sand F65 granular media under different rotational and vertical velocities are shown in figure 3.5 and figure 3.6. Penetration force decreases with the increase of rotational speed. Further, it decreases due to the decrease of vertical velocities during the clockwise penetration test, as shown in figure 3.5 (a). In this preliminary study, from figure 3.5 (a), it is obvious that after

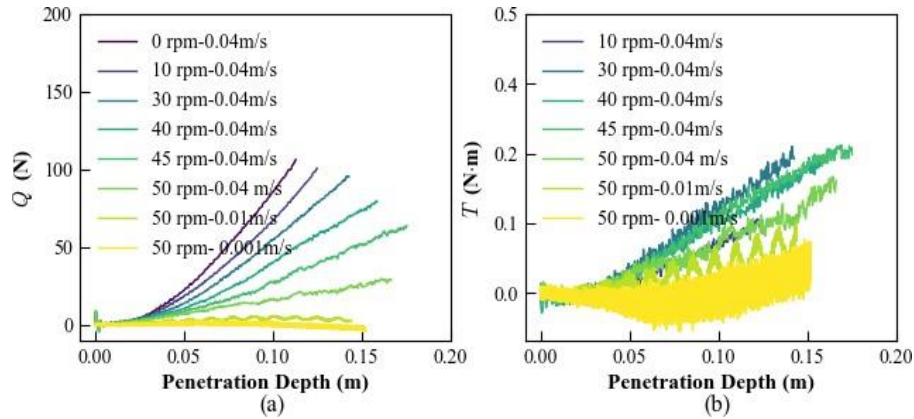

**Figure 3.7:** Clockwise Penetration Force and Torque for Right Handed Helical Penetrator Under Different Rotational Velocities in Glass Beads: (a) Penetration Force ($Q$); (b) Penetration Torque ($T$).

50 rpm-0.04 m/s, the granular media is under a critical state and due to further decrement of vertical velocities negative force is found which will be more convenient for burrowing. To better understand the critical state,, further investigation will be done. However, penetration torque also decreases concerning increasing rotational speed, which is supposed to be not expected. This can happen due to the small inclination angle of the helical penetrator. More research is needed to know the actual reason for torque data for the clockwise helical penetrator. The penetration force decreases with the increase of rotational speed during counterclockwise penetration for 0.04 m/s vertical velocities, as shown in figure 3.6 (a), but it further starts increasing for 50 rpm-0.001 m/s state because during counterclockwise penetration particle started to accumulate downward as shown in the figure 3.3 (c) which generate excessive force. However, penetration torque increases with the increase of rotational velocities, as shown in figure 3.6 (b). This observation implies that clockwise penetration is easier for burrowing sand than counterclockwise penetration.

### 3.3.2  Clockwise and Counterclockwise Downward Penetration in Glass Beads

The penetration force and torque for glass beads under different rotational speeds of helical penetrator are shown in figure 3.7 and figure 3.8. Penetration force decreases for increasing rotational speed and decreases for decrements of vertical velocities during clockwise penetration tests as shown in figure 3.7 (a). The reason for further decrements is that during clockwise penetration, particles come out, as shown in figure 3.3 (b). Penetration torque increases for the increment of rotational speed, and it decreases due to decrements of vertical velocities for clockwise penetration as shown in figure 3.7 (b). During counterclockwise penetration tests, for glass beads, penetration force and torque do not change that much due to the increment of rotational speeds, as shown in figure 3.8. The main reason behind this phenomenon is that particles accumulate downward during counterclockwise penetration, and sharing blades of helical penetrators get almost the same amount of glass beads particles during penetration.

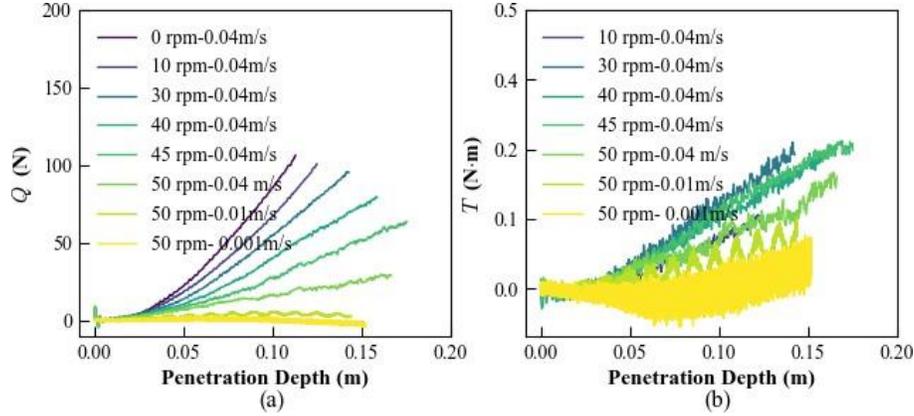

**Figure 3.8:** Counterclockwise Penetration Force and Torque for Right Handed Helical Penetrator Under Different Rotational Velocities in Glass Beads: (a) Penetration Force ($Q$); (b) Penetration Torque ($T$).

### 3.3.3 Surface Topography For Clockwise and Counterclockwise Penetration

Image processing was performed to reconstruct the experiment scenario and demonstrate the surface topography. About 100 pictures were taken from different angles and orientations for each case. The pixel points were processed using the open-access software VisualSFM to reconstruct a 3D structure consisting of plenty of mesh points. Software Cloudcompare was then employed to process and align the dimensions of the mesh points.

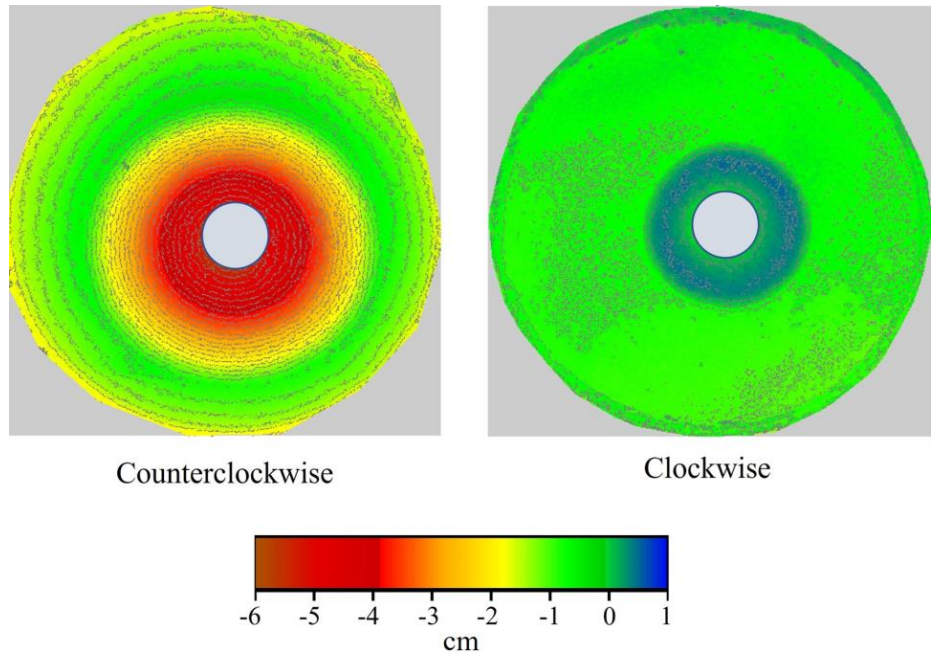

**Figure 3.9:** Surface Topography in Different Penetration Manner.

Figure 3.9 shows the surface topography from different rotation manners using the above method.

### 3.3.4 *Normalized Clockwise and Counterclockwise Penetration Force Under Different Rotational Modes*

The normalized clockwise and counterclockwise penetration force under different rotational modes are shown in figure 3.10. It is observed that normalized penetration force decreases with the increase of relative slip velocity. The term relative slip velocity is defined as the ratio between the velocity of rotational movement and the velocity of vertical penetration. The result confirms that clockwise rotational penetration forces approach around 1% for relative slip velocity is 133 for Ottawa sand and glass beads. In the counterclockwise state for glass beads and Ottawa sand, penetration force approaches 32% compared with a relative velocity is 133. Based on these observations, the rotational force decrement rate is higher for clockwise penetration than counterclockwise penetration. Here are the lessons and findings from the vertical penetration tests as shown in the table 5.1.

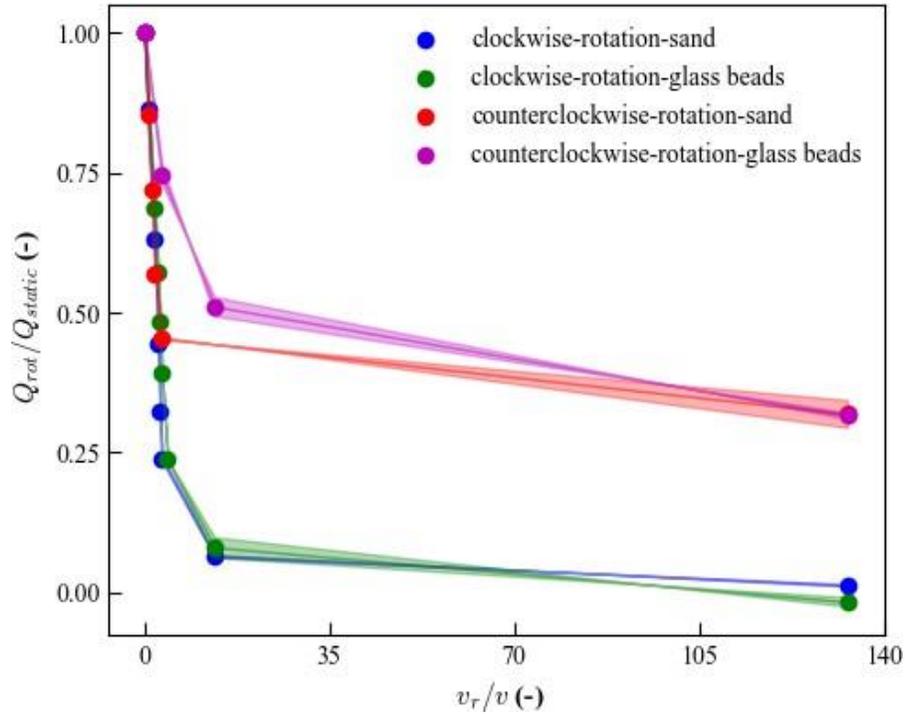

**Figure 3.10:** The normalized Penetration Force $Q_{rot}/Q_{static}$ for Different Rotational Modes.

### 3.3.5   Sliding Test Result

We conducted a sliding testing process to investigate the possibility of achieving translation through pure rotation. The test involved a slider and a rotational auger in a dense sample with a relative density of 88%. The results showed that during the clockwise rotation of a right-handed auger, the net thrust force acted downward, facilitating downward penetration as shown in figure 3.11. Conversely, during counterclockwise rotation or upward burrowing of a right-handed auger, the net thrust force acted upward, aiding in burrowing out as shown in figure 3.11. Thus, we can conclude that the combination of self-weight and rotation allows the auger to penetrate the soil without requiring external force.

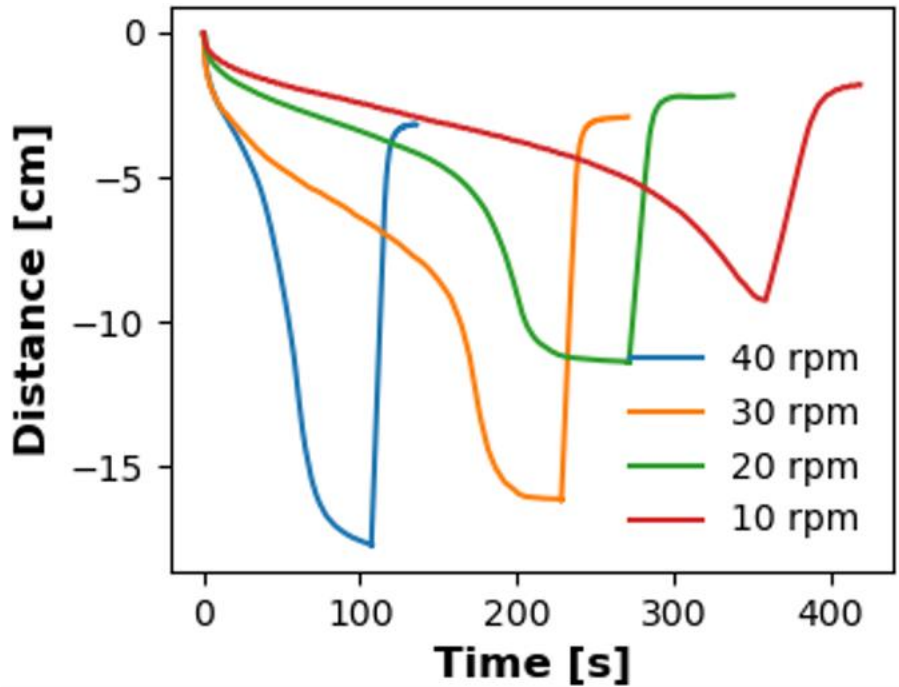

**Figure 3.11:** Sliding Test Graph for Clockwise and Counterclockwise Penetration Under Different Rotation Modes

## 3.4   CONCLUSION

In this paper, a series of downward rotational penetration tests are conducted using a helical penetrator to study the effect of penetrator geometry & different granular media. The testing setup consists of a universal robot and a motorized helical penetrator. Highlighting the effect of asymmetric kinematics, this paper shows that breaking symmetry is much easier than counterclockwise penetration during clockwise penetration. This study affirms that reducing the anisotropy of friction and symmetry breakdown using the helical shape penetrator determines locomotion in granular media. This result gives us the insight to design a burrowing robot based on the result of the experiment.

Chapter 4

DUAL-AUGER HORIZONTAL BURROWING ROBOT

4.1  INTRODUCTION

Bio-inspired design helps to solve many engineering problems. Many mobile robots are studied that are available on land, water and in space, but research in the locomotion of robots through subterranean is not yet well investigated (Aguilar *et al.*, 2016). The difficulties include complex soil behavior, and the unsaturated and dense environment of granular media. In recent years scientists are showing great interest in designing a robot to operate in granular media, using different burrowing strategies. Various biological animals (Maladen *et al.*, 2009; Jung, 2010; Winter *et al.*, 2012) and plant roots (Bengough and Mullins, 1990; Bengough *et al.*, 2008) locomote through the granular media by significantly reducing intergranular resistance. *Erodium* and *Pelargonium* flowering plants can bury into the granular media through hygroscopic coiling and uncoiling motions of their awns. The helical structure helps to break the symmetry of granular media. Inspired by the burrowing mechanisms of *Erodium seeds* for rotation and *Scincus scincus* for helical motion, a horizontal burrowing robot is designed, which is featured by two augers and is stabilized by a fin structure in the middle. The fin works as an anchor and helps the burrowing robot travel through the medium. The robot is buried 10cm below the surface of glass beads, and self-burrowing tests, direct drag, rotational drag tests and thrust tests at various rotational speeds explore the burrowing mechanism. Finally, to achieve the self-burrowing, we tested our hypothesis thrust force should be greater than or equal to the drag force.

Here is the table 4.1 showing the testing scenario to understand the burrowing mechanisms of a self-burrowing robot :

**Table 4.1:** Tasks for Investigating Burrowing Characteristics and Force Balances

| Objective | Tasks |
| --- | --- |
| Investigate burrowing characteristics | Horizontal and vertical burrowing tests |
| Force balances and force comparisons | - Thrust Test <br> - Drag Test (Direct Drag, Rotational Drag) |
| Hypothesis | Thrust ≥ Drag |

## 4.2 METHODOLOGY

### 4.2.1 Dual-auger Robot Design

Generally, biological organisms use internal forces and body deformation to facilitate underground burrowing. Biological organisms know how to coordinate the formation of different parts of the body to generate thrust to overcome backward slip and facilitate forward movement. Underground movement is considered a soil-organism interaction problem compared to soil-structure interaction problems. It has been observed that effective underground locomotion is related to different burrowing strategies and characteristics of the surrounding soil. In this study, a self-burrowing robot is designed that can move horizontally in a glass beads pool. A series of burrowing tests are conducted under different conditions. It is found that translations can be happened by breaking symmetry or boundary conditions in properties of granular medium (Maladen *et al.*, 2011b; Tao *et al.*, 2020). So in order to achieve horizontal burrowing symmetry, breaking features should be introduced. It is observed that rotation can reduce penetration resistance to move forward (Tang and Julian Tao, 2021; Tang *et al.*, 2020b). The combination of rotational and propulsive motion along the helical axis of two augers helps break the kinematic symmetry of granular materials, reducing penetration resistance.

In this study, we propose a self-burrowing robot that consists of two major parts:

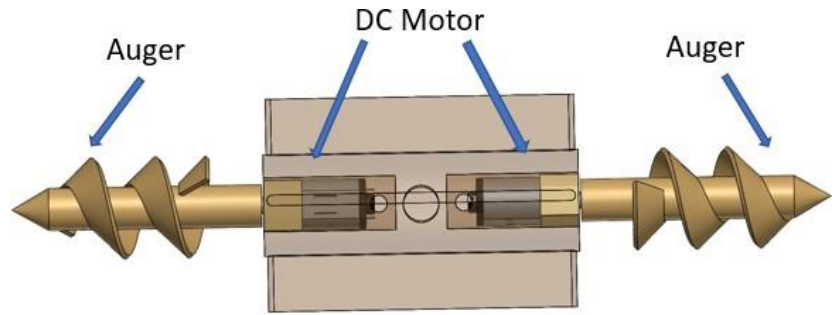

**Figure 4.1:** Dual-auger Robot Design

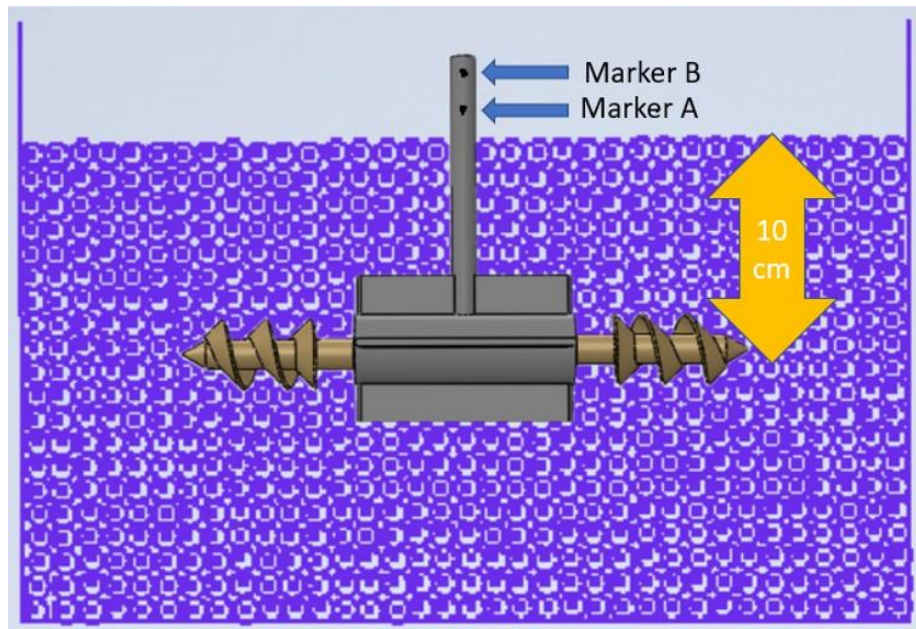

**Figure 4.2:** Illustration of Self-burrowing Testing Setup

an Auger and four fin-structured stators. The dimension of each 3D printed auger is (24mmx24mmx60mm) as shown in figure 4.1. The dimension of a 3D printed stabilized fin rotor is (50mmx50mmx70mm) 4.1. The rotation of the auger is controlled by a DC gear motor, and the highest rpm is 220.

### *4.2.2 Self-burrowing Test Set Up*

A testing setup is created to evaluate the performance of the self-burrowing robot in a pool of glass beads as shown in figure 4.2. The experimental setup consists of a glass bead container(dimension: 60cm * 18cm * 30cm), a camera to record the movement of the robot and a horizontal two-auger robot buried in the glass beads at a depth of 10 cm. A steel mast(31cm) is fixed in the middle of two auger robots positioned perpendicular to the glass bead container's bottom. The two auger robot is buried in the middle of the glass bead container. The steel mast top portion extends beyond the T-slot framework track, which restricts the mast movement to the glass bead container in the longitudinal direction. The marker's real-time position and the robot's inclination is monitored via video tracking the marker A point (11.5 cm above the robot) and B point(13.5 cm above the robot). The open-source computer-vision library OpenCV is used to process the obtained videos of burrowing robot (Minichino and Howse, 2015). The optical flow algorithm based on the Lucas-Kanade method (Lucas and Kanade, 1981) is employed to extricate the trajectory of the marker situated on the mast from the video. The performance of the two auger self-burrowing robot is then defined by the moving characteristics of marker A as shown in figure 4.2.

### 4.2.3  Drag, Thrust Testing Set Up

Direct drag, rotational, and thrust tests are introduced to further predict the burrowing behavior of the two auger self-burrowing robots. The thrust test helps us understand how much force the robot requires to move forward. The full experimental setup consists of a UR16e universal robot, two auger robots, a 3D printer coupler, and a steel rod, as shown in figure 4.3. The two auger robot lower part is connected with a steel rod (Diameter 8mm, length 40cm), and the steel rod upper part is connected with a 3D printed coupler (9cm X 9cm X 1.5 cm) to a universal robotic arm. The six-axis force and torque sensor systems integrated into the UR16e universal robot enable real-time measurement of the penetration force and torque during the process. The universal robot is capable of capturing mechanical loads and converting them into forces and torques in all directions. To record the data on penetration force and torque, which was obtained from the control box of the robotic arm, the Real-Time Data Exchange python package was utilized to transmit it to the local drive. However, to keep the density of the samples accurate and precise, dry pluviation preparation technique is applied (Okamoto and Fityus, 2017). According to laboratory measurements, the actual density of granular media is Dr=46.2%. For the thrust test setup, the robot will be in the stationary position, and the upper part is connected with universal robot UR16e, and the two augers will be rotating. The generated thrust is recorded from the universal robot for 210 rpm, 160 rpm and 105 rpm. Direct Drag force is generated by the stator and augers in a pure translational movement. The rotational drag test is introduced to measure the drag when a robot is moving through the glass beads pool. Two auger robots move horizontally under different rotational and horizontal velocities during the rotational drag test. This net force on the augers is believed to consist of both drag and thrust forces. The horizontal velocities of the dual-auger robot are controlled by UR16e universal robot. The rotational movement of the DC gear motor is controlled by an ATmega2560 microcontroller. Arduino ATmega 2560 provides a PWM(Pulse width modulation) signal to the motor driver, and the motor driver follows the microcontroller's command and controls the DC motor's speed.

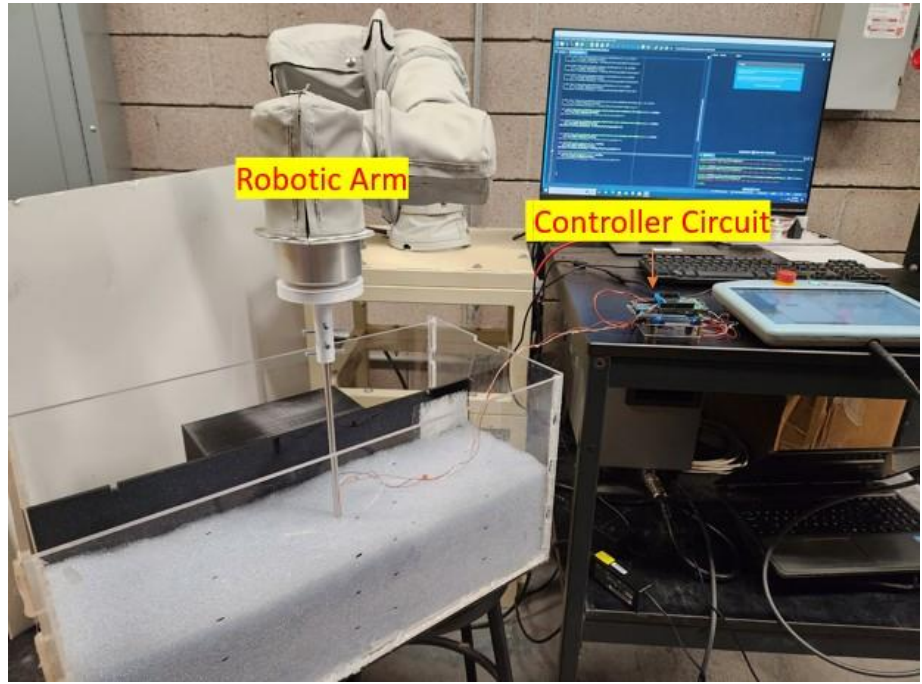

**Figure 4.3:** Experimental Testing Setup for Direct Drag, Rotational Drag, and Thrust Test

### 4.3 RESULT and DISCUSSION

#### 4.3.1 *Burrowing Characteristics*

Figure 4.4 illustrates the burrowing behavior of the robot in the horizontal direction for three different rpm 210, 160 and 105. By applying rotational and propulsive movement along the helical axis of two augers, the kinematic symmetry of granular materials is disrupted, leading to a decrease in penetration resistance. It is observed from the first 30 seconds of burrowing characteristics for all three different rpm are almost linear, and we can predict the burrowing speed of the robot, which is around 0.08 mm/s. After the first 30 seconds, due to tilting or uplift force, it moves closer to the surface. The travel distance of the robot in a horizontal direction is around 20 centimetres.

Figure 4.5 illustrates the burrowing behavior of dual-auger robots in the vertical direction for three different rpm 210, 160 and 105. Due to uplift force, the robot moves vertically. The robot moves around 5cm distance during the vertical movement.

Figure 4.6 illustrate the inclination angle of the robot. The inclination of the robot in the vertical plane is determined by calculating the angle between the axial direction of the mast and the surface of the ground. The estimation of the inclination angle is based on the extracted positions of two markers, namely Marker A and Marker B. Here we can observe from the figure 4.6 it starts from a 90-degree angle and have a uniform or linear part in the begging, and then it accelerates when it approaches the surface.

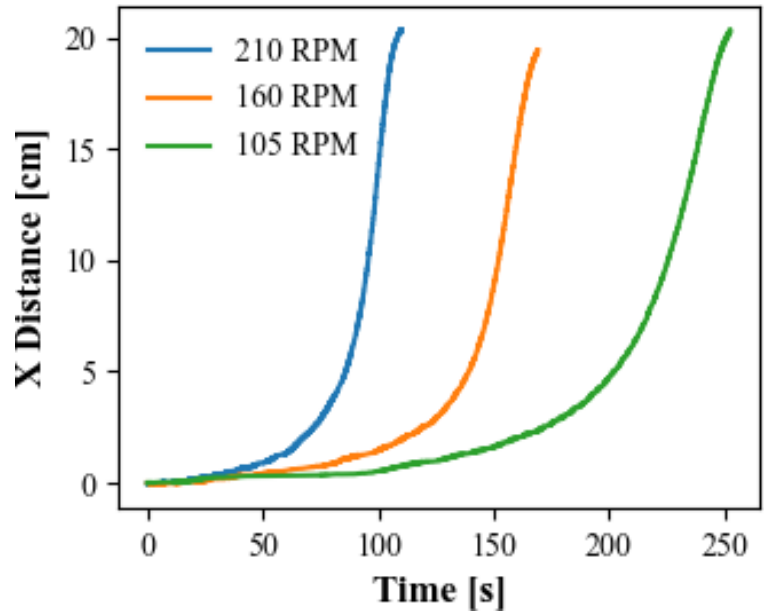

**Figure 4.4:** The Horizontal Movement of the Dual-auger Burrowing Robot in Glass Beads

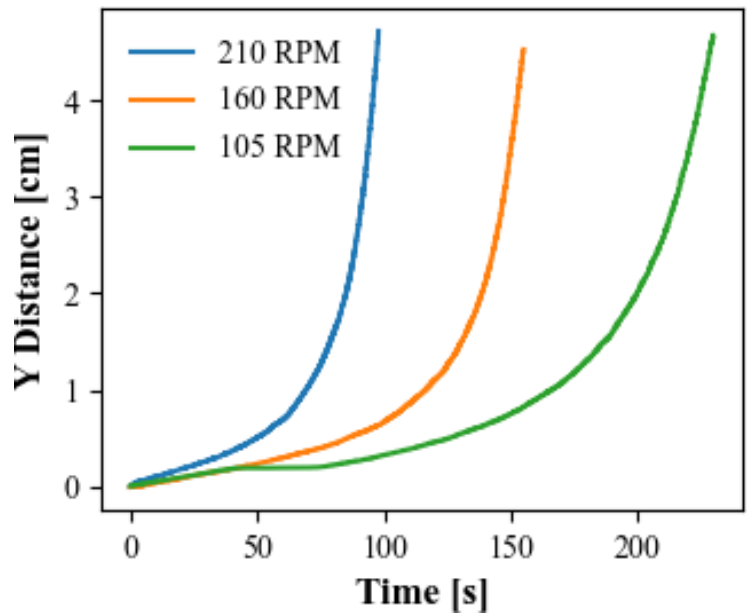

**Figure 4.5:** The Vertical Movement of the Dual-auger Burrowing Robot in Glass Beads

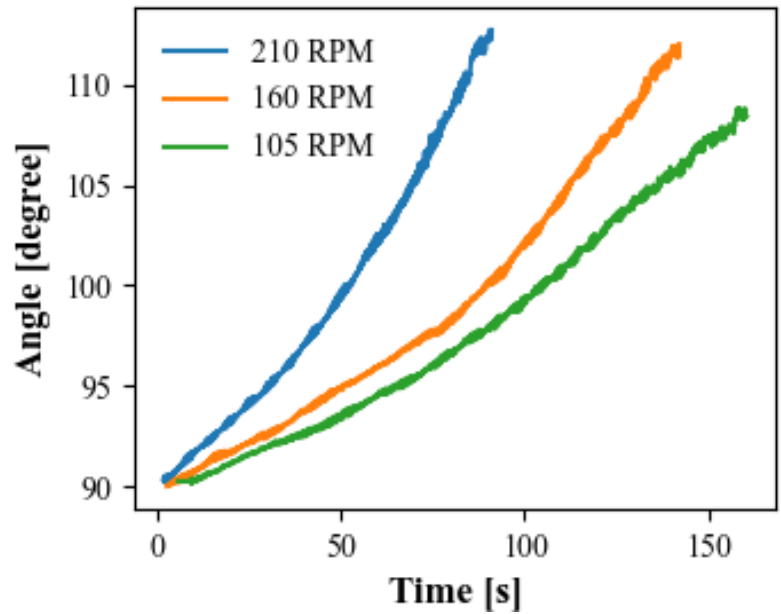

**Figure 4.6:** Inclination Angle of the Dual-auger Burrowing Robots

### 4.3.2 Direct Drag

We conducted a study to investigate the effect of horizontal velocity on drag without auger rotation. Two different horizontal velocities(10 mm/s, 5 mm/s) were selected and tested for their effect on direct drag. The results showed no effect of horizontal velocity during the pure drag test, as depicted in figure 4.7.

### 4.3.3 Rotational Drag

We further investigate the effect of rotation on drag resistance. Figures 4.9 and 4.10 show that for 210 rpm and horizontal velocities of 10 mm/s and 5 mm/s, the drag force is significantly reduced compared to the direct drag case shown in figure 4.7. Table 4.2 provides a comparison of drag forces and rotational drag forces for different relative slip velocities and rotational speeds.

Now we want to investigate the effect of relative slip velocity for different rotational drags. Figure 4.8 displays the rotational drag force for a relative slip velocity of 10. The graph reveals that for the same relative slip velocity, the rotational drag force remains constant regardless of the rotational and horizontal velocities. The graph contains three different curves, each representing a different combination of rotational and horizontal velocities with the same ratio. Interestingly, the calculated drag force is approximately 18 N for all three curves.

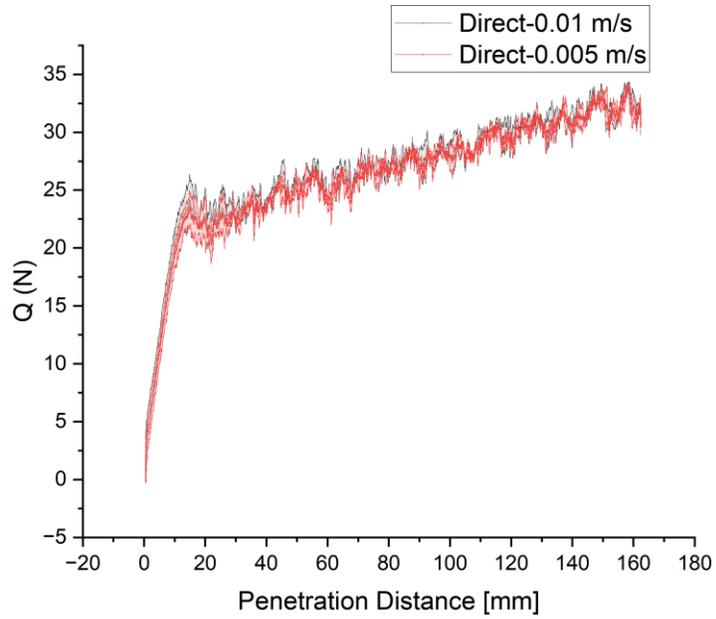

**Figure 4.7:** Direct Horizontal Drag of Dual-auger Robots

**Table 4.2:** Drag Forces for Different Test Cases

| Test Case | Horizontal Velocity (mm/s) | Drag Force (N) |
|---|---|---|
| Direct Drag(Without rotation) | 10 | 30 |
|  | 5 | 30 |
| Rotational Drag(210 rpm) | 10 | 12 |
|  | 5 | 10 |

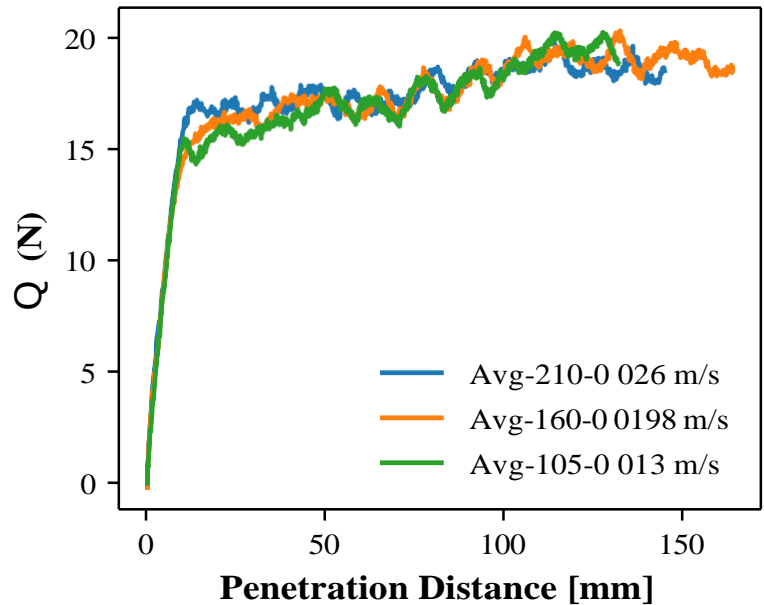

**Figure 4.8:** Relative Slip Velocity 10 : Right Handed Clockwise Horizontal Burrowing Forces(N) Under Different Rotational and Horizontal Velocities in Glass Beads ($D = 2.0$ mm).

Figure 4.9 displays the rotational drag force for a relative slip velocity of 26. The graph reveals that for the same relative slip velocity, the rotational drag force remains constant regardless of the rotational and horizontal velocities. The graph contains three different curves, each representing a different combination of rotational and horizontal velocities with the same ratio. Interestingly, the calculated drag force is approximately 12.5 N for all three curves.

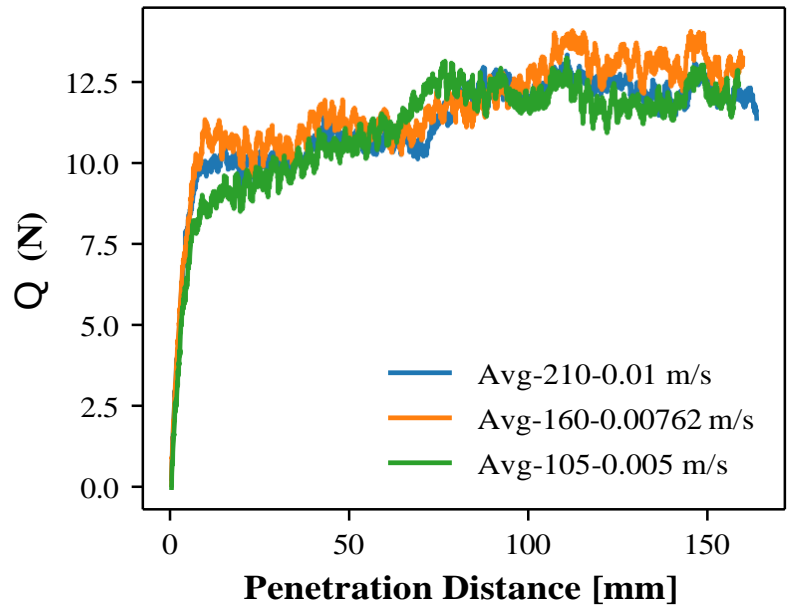

**Figure 4.9:** Relative Slip Velocity 26 : Right Handed Clockwise Horizontal Burrowing Forces(N) Under Different Rotational and Horizontal Velocities in Glass Beads ($D = 2.0$ mm).

Figure 4.10 displays the rotational drag force for a relative slip velocity of 53. The graph reveals that for the same relative slip velocity, the rotational drag force remains constant regardless of the rotational and horizontal velocities. The graph contains three different curves, each representing a different combination of rotational and horizontal velocities with the same ratio. Interestingly, the calculated drag force is approximately 10 N for all three curves.

Figure 4.11 displays the rotational drag force for a relative slip velocity of 132. The graph reveals that for the same relative slip velocity, the rotational drag force remains constant regardless of the rotational and horizontal velocities. The graph contains three different curves, each representing a different combination of rotational and horizontal velocities with the same ratio. Interestingly, the calculated drag force is approximately 8.5 N for all three curves.

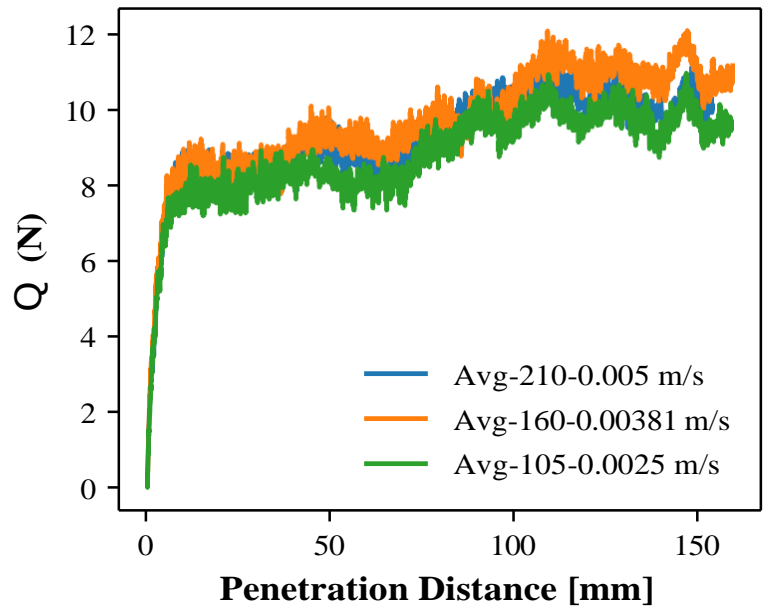

**Figure 4.10:** Relative Slip Velocity 53 : Right Handed Clockwise Horizontal Burrowing Forces(N) Under Different Rotational and Horizontal Velocities in Glass Beads ($D = 2.0$ mm).

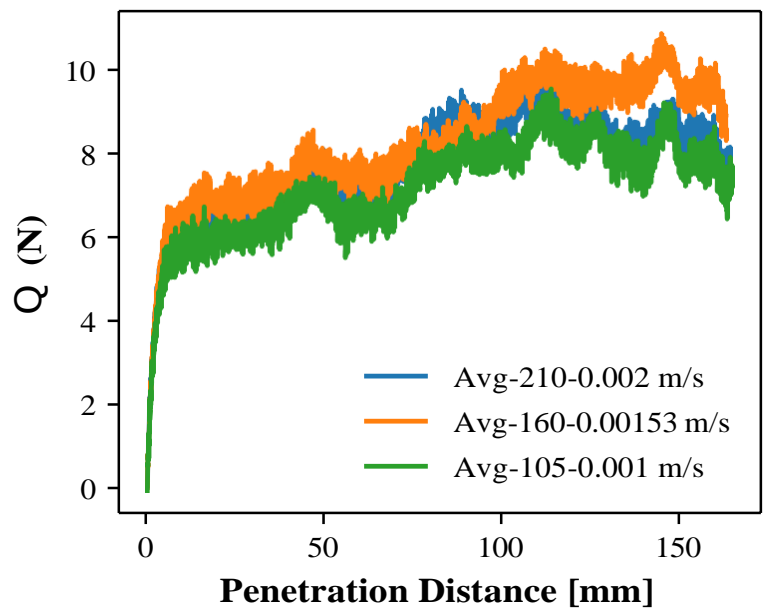

**Figure 4.11:** Relative Slip Velocity 132: Right Handed Clockwise Horizontal Burrowing Forces(N) Under Different Rotational and Horizontal Velocities in Glass Beads ($D = 2.0$ mm).

**Table 4.3:** Calculated Rotational Drag Forces for Different Relative Slip Velocities

| Relative Slip Velocity | Rotational Speed | Horizontal Speed | Rotational Drag Force |
|:---:|:---:|:---:|:---:|
| 10 | 210 rpm | 26 mm/s | 18 N |
|  | 160 rpm | 19.8 mm/s |  |
|  | 105 rpm | 13 mm/s |  |
| 26 | 210 rpm | 10 mm/s | 12.5 N |
|  | 160 rpm | 7.62 mm/s |  |
|  | 105 rpm | 5 mm/s |  |
| 53 | 210 rpm | 5 mm/s | 10 N |
|  | 160 rpm | 3.81 mm/s |  |
|  | 105 rpm | 2.5 mm/s |  |
| 132 | 210 rpm | 2 mm/s | 8.5 N |
|  | 160 rpm | 1.53 mm/s |  |
|  | 105 rpm | 1 mm/s |  |

From table 4.3, it is obvious that for the same relative slip velocity, the rotational drag force is the same for different rotational and horizontal velocities.

We want to check the relationship between normalized rotational drag force and relative slip velocity. Our findings suggest that normalized rotational drag force decreases with the increase of relative slip velocity, as shown in figure 4.12. Specifically, for a relative slip velocity of 10, the rotational drag force reduces to 60%; for 26, it reduces to 45%; for 53, it reduces to 30%; and for 132 it reduces to 27%. These results demonstrate an inverse relationship between the normalized rotational drag force and the relative slip velocity.

We investigated the effect of auger design to change the drag forces. From the figure 4.13, it is found that for 210 rpm and 5 mm/s horizontal velocity double helix auger design calculated rotational drag force is around 14 N, and for 210 rpm and 5 mm/s horizontal

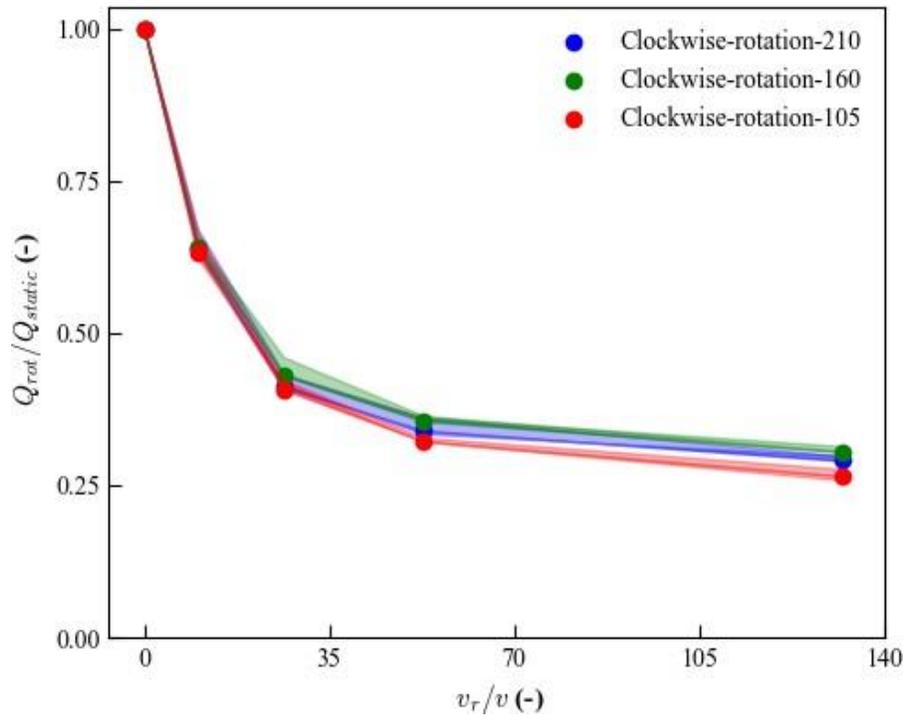

**Figure 4.12:** Normalized Rotational-drag Force $Q_{rot}/Q_{static}$ for Right Handed Clockwise Rotation.

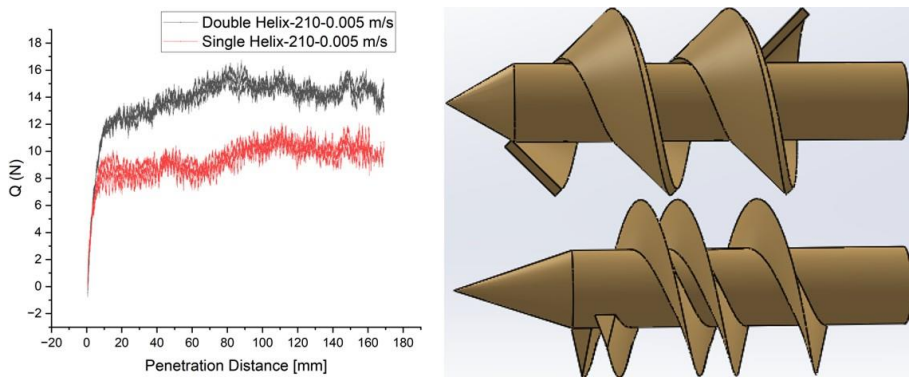

**Figure 4.13:** Single Helix and Double Helix Auger Effect to Drag Force

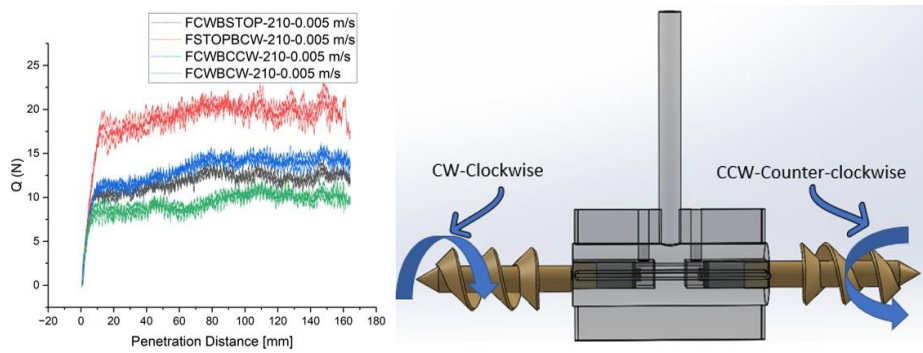

**Figure 4.14:** Single Helix Dual-Auger Robots Different Kinematics Effect to Drag Forces

velocity single helix auger design calculated rotational drag force is around 14 N.

We conducted a further investigation to analyze how the kinematics of the augers affect drag forces. The results are presented in figure 4.14. For a fixed horizontal velocity of 5 mm/s and a rotational speed of 210 rpm, we tested three cases: front auger right-handed clockwise rotation and back auger STOP, front auger right-handed clockwise rotation and back auger right-handed counter-clockwise rotation, and front auger STOP and back auger right-handed clockwise rotation. We found that the drag force was almost the same for the first two cases, around 12 N, as shown in figure 4.14. However, for the case where the front auger was STOP, and the back auger was right-handed clockwise rotation, the calculated drag force was around 20 N. Similarly, for the case where both the front and back augers were right-handed clockwise rotation, the calculated drag force was around 9 N. Therefore, we can conclude that different auger kinematics lead to different calculated drag forces.

### 4.3.4 Thrust Test

The augers generate thrust force through pure rotational movement. The thrust test graph for the horizontal burrowing robot is shown in Figure 4.15 for 210 rpm, 160 rpm, and 105 rpm. Thrust force is measured while the robot is stationary and the augers are rotating. The thrust test values are approximately 3.8 N for 210 rpm, 3.4 N for 160 rpm, and 1.6 N for 105 rpm, as shown in Figure 4.15. The thrust test helps us to evaluate the robot's performance in moving forward. Figure 4.15 shows that thrust force increases with increasing rotational speed.

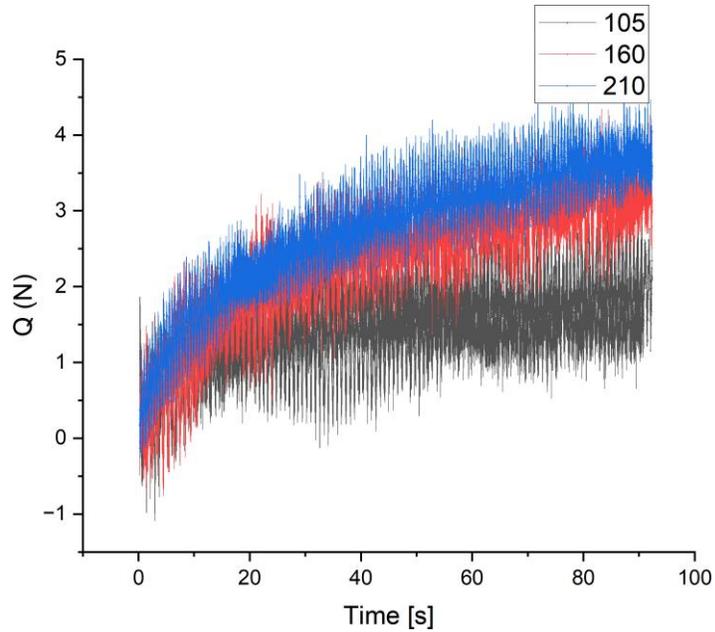

**Figure 4.15:** Thrust Test Data for 210 rpm, 160 rpm and 105 rpm.

### 4.3.5 Self-burrowing Hypothesis

To verify our hypothesis that the thrust should be higher than or equal to the drag for enabling self-burrowing, we conducted tests at a fixed horizontal velocity of 0.1 mm/s using a universal robot. The actual velocity of the robot during horizontal burrowing was measured at 0.08 mm/s, as shown in figure 4.4. To achieve almost the same horizontal velocity, we fixed the horizontal velocity of the universal robot at 0.1 mm/s and tested the robot for 210 rpm and 160 rpm rotational velocities. As shown in figure 4.16 (a), for the direct drag of the rod, the drag force was found to be approximately 6.5 N. The rotational drag force at 210 rpm and 0.1 mm/s horizontal velocity was approximately 8.2 N, while the rotational drag force at 160 rpm and 0.1 mm/s horizontal velocity was around 8.5 N.

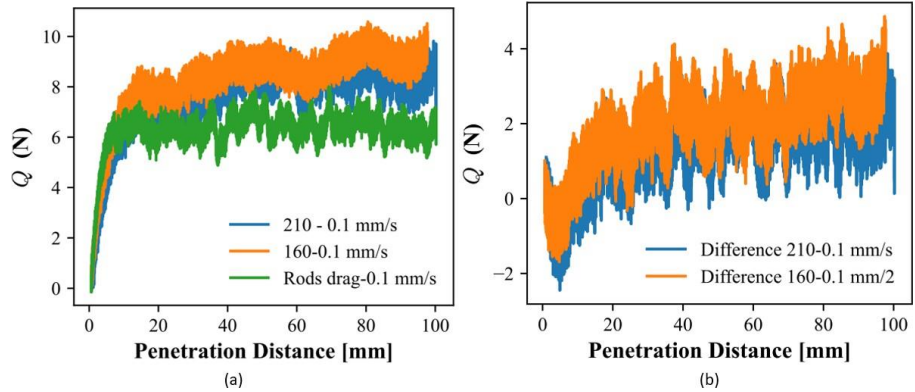

**Figure 4.16:** Self-burrowing Hypothesis Illustration: (a) Rotational Drag for 210 rpm and 160 rpm for Single Helix and Only Rod Direct Drag. (b) Calculated (Robot Drag+ Thrust) from Rotational Drag and Rod.

To obtain the net thrust force(Robot Drag + Thrust) acting on the robot during burrowing, we subtracted the direct drag force of the rod from the rotational drag force at 210 rpm and 160 rpm with a fixed horizontal velocity of 0.1 mm/s. Figure 4.16 (b) illustrates that the net thrust force is approximately 3.1 N and 3.3 N for 210 rpm and 160 rpm, respectively. On the other hand, the calculated actual thrust force for 210 rpm is 3.8 N and for 160 rpm is 3.4 N, as shown in figure 4.15. Therefore, we can conclude that the thrust force is higher than the drag force based on this relationship.

## 4.4 CONCLUSION

In conclusion, bio-inspiration is a valuable approach to engineering design as it allows us to take inspiration from nature and apply those principles to develop innovative engineering solutions. One such example is the development of a self-burrowing horizontal robot with a dual-auger design.

The dual-auger design of the robot facilitates horizontal movement, allowing it to navigate through granular media efficiently. The results of the experiments conducted on the robot showed that the horizontal self-burrowing of the robot increased with increasing rotational speed. This is an important finding as it can help engineers optimise similar robot design for maximum efficiency.

Another important conclusion drawn from the experiments is that for the robot to enable self-burrowing, the thrust generated by the augers must be greater than or equal to the drag force experienced by the robot. This finding highlights the importance of understanding the balance between these two forces when designing self-burrowing robots.

Overall, the development of this self-burrowing horizontal robot with a dual-auger design is a promising advancement in the field of robotics. The use of bio-inspiration in the design process allows for the development of robots that can easily navigate complex environments. The experiments' results on the robot provide valuable insights into the relationship between rotational speed, thrust, and drag, which can inform future design decisions.

One potential application for this self-burrowing robot is in the field of agriculture. The ability of the robot to move horizontally through soil can be useful for tasks such as planting or fertilizing crops.

In addition, the insights gained from this research can also be applied to the development of other types of self-burrowing robots, such as those designed for subterranean exploration or underground construction. By understanding the balance between thrust and drag, engineers can optimize the design of these robots for maximum efficiency and effectiveness.

In conclusion, developing a self-burrowing horizontal robot with a dual-auger design demonstrates the potential of bio-inspiration in engineering design. The experiments' results on the robot provide valuable insights into the relationship between rotational speed, thrust, and drag, which can inform future design decisions for similar robots. The potential applications of this technology in geotechnical site investigation, subterranean exploration, and precision agriculture make it a highly promising and Interdisciplinary field of research for the future.

Chapter 5

SUMMARY, CONCLUSIONS AND FUTURE WORK

Motivated by exceptional burrowing behavior of self-burrowing rotary seeds, *Erodium cicutarium*, and undulation motion of *Scincus scincus*, a series of downward penetration tests and a dual auger self-burrowing robot is reported in this thesis study. *Erodium cicutarium* can dig into the soil for future germination using a continuous rotational motion of the awn. The awn's bi-layered structure facilitates it to produce varying deformations in response to changes in humidity, leading to a coiling and uncoiling motion. The awn rotates during extension due to its helical form and hygroscopic expansion effects.

Its motion is similar to a traditional auger or drilling machine with a spinning intruder. As a result, the seeds are believed to spin themselves to help them penetrate the soil. However, more research needs to be made investigating the burrowing behavior of the *Erodium cicutarium* and the undulation motion of *Scincus scincus*. The impact of a helical rotating auger in various granular media is investigated through a series of downward rotational penetration tests using a helical penetrator. A motorized helical penetrator and a universal robot complete the downward penetration. According to this study, the helical form penetrator determines locomotion in granular media by minimizing friction and symmetry breakdown anisotropy.

The below table 5.1 shows the lessons learned from downward penetration tests :

To further investigate the horizontal burrowing behavior, a dual-auger self-burrowing robot is designed and tested for self-burrowing characteristics, thrust and rotational drag. Modeling the self-burrowing robot's burrowing behavior can be investigated in future to predict that thrust force should be higher than drag force.

In order to further improve the design and capabilities of the dual-auger self-burrowing

**Table 5.1:** Lessons Learned From Downward Penetration Tests

| Lesson | Findings |
|--------|----------|
| 1 | For the same vertical velocity, the resistive force decreases with the increase of the rotational speed |
| 2 | For the same rotational velocity, resistive force further decreases with the decrease of the vertical velocity |
| 3 | Overall, the dominant factor is actually the relative slip velocity which describes the relative contribution of the rotational speed and vertical speed |
| 4 | Clockwise rotation results in lower resistance and requires lower force than counterclockwise |

robot, several areas of future work could be explored.

Firstly, numerical simulations could be conducted to model the robot's behavior and compare the results with experimental data. This will help us to get a better understanding of the robot burrowing dynamics.

Secondly, the auger design could be optimized to increase thrust and efficiency. This could involve tweaking the pitch and angle of the augers for improved durability and performance.

Lastly, the robot could be tested in different types of soils and at varying depths to assess its versatility and potential for use in various applications. This could involve field testing in real-world scenarios or controlled laboratory experiments to evaluate its performance under different conditions.

By conducting these future works, we can enhance the efficiency and effectiveness of the dual-auger self-burrowing robot, making it a more valuable tool for geotechnical site investigation, subterranean exploration, and precision agriculture, among other potential applications.